\documentclass[runningheads]{llncs}

\usepackage{bbding}
\usepackage{amssymb}
\usepackage{array}
\usepackage{graphicx}
\usepackage{bm}
\usepackage{graphicx}
\usepackage[colorlinks,linkcolor=blue,anchorcolor=blue,citecolor=blue]{hyperref}
\begin{document}
\title{Dual Branch Network Towards Accurate Printed Mathematical Expression Recognition}

%
\titlerunning{Dual Branch Network Towards Accurate PMER.}
%
\author{Yuqing Wang\inst{1}\orcidID{0000-0001-6384-1753} \and
Zhenyu Weng\inst{1}\orcidID{0000-0001-7857-8687} \and
Zhaokun Zhou\inst{1}\and
Shuaijian Ji\inst{1}\and
Zhongjie Ye\inst{1}\and
Yuesheng Zhu\inst{1}\orcidID{0000-0003-2524-6800}}
\authorrunning{Y. Wang et al.}
%
\institute{Communication and Information Security Lab, Shenzhen Graduate School ,
Peking University, Shenzhen, China\\
\email{\{wyq, zhouzhaokun, 2101212809, zhongjieye\}@stu.pku.edu.cn,\\ \{wzytumbler, zhuys\}@pku.edu.cn}\\
}
\maketitle              
\begin{abstract}
    Over the past years, Printed Mathematical Expression Recognition (PMER) has progressed rapidly. However, due to the insufﬁcient context information captured by Convolutional Neural Networks, some mathematical symbols might be incorrectly recognized or missed. 
    To tackle this problem, in this paper, a Dual Branch transformer-based Network (DBN) is proposed to learn both local and global context information for accurate PMER. In our DBN, local and global features are extracted simultaneously, and a Context Coupling Module (CCM) is developed to complement the features between the global and local contexts. 
    CCM adopts an interactive manner so that the coupled context clues are highly correlated to each expression symbol. Additionally, we design a Dynamic Soft Target (DST) strategy to utilize the similarities among symbol categories for reasonable label generation. Our experimental results have demonstrated that DBN can accurately recognize mathematical expressions and has achieved state-of-the-art performance.

\end{abstract}
\section{Introduction}

Mathematics expression understanding has received much attention from both academia and industry due to its numerous applications such as arithmetical exercise correction~\cite{pigai}, student performance prediction~\cite{student}, and automatic marking~\cite{automarket}. Deep neural networks have made dramatic advances over the past few years, resulting in many inspiring ideas \cite{d3,d4,edsl,yu2022continual,yu2022multi,yu2023metamath}. However, there is still large room for improvement when facing expressions containing semantically-correlated symbols.

\begin{figure}[t]
    \centering
    \includegraphics[width=0.8\textwidth]{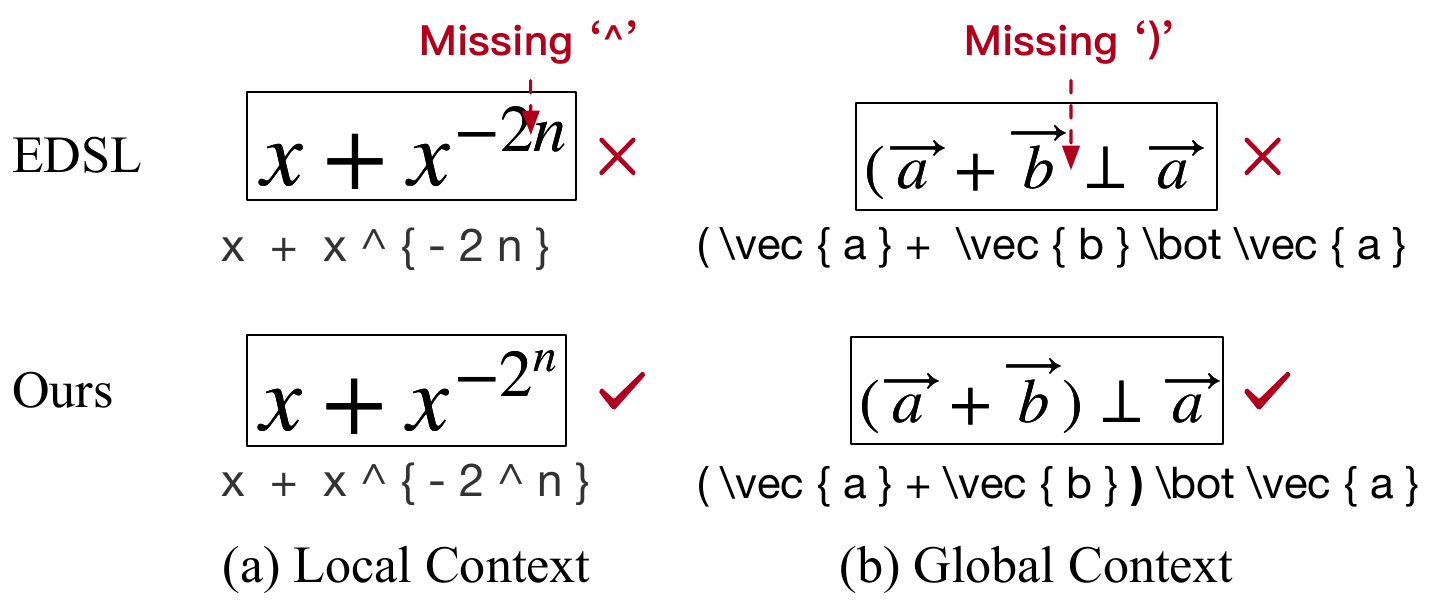}
    \caption{The predicted LaTeX sequence from EDSL and our methods. We also rendered the predicted LaTeX sequence into images for better visualization. EDSL made some missing predictions while
    ours predictions are the correct results.}
    \label{page1}
  \end{figure}

  Mathematics expression recognition takes a mathematical expression image as input and transforms the image into the format of MathML or LaTeX sequence. Due to the inherent complicated semantic correlation of the mathematical expression, learning the context information between symbols plays a crucial role in PMER. However, the context information captured by existing methods is still insufficient, which may result in suboptimal predictions. For example, EDSL~\cite{edsl}, a representative transformer-based method which treats segmented symbols as image patches, will degrade when recognizing expressions containing semantically correlated symbols and result in incorrect results as depicted in Fig.~\ref{page1}. We argue that the reason is the insufficient context information which comes from two aspects: {\bfseries 1). Insufficient local context clues.} The symbol that determines the local spatial structure is highly related to its neighboring symbols, for instance, the exponent symbol `\^{}' between the base number `2' and the exponent `n', as depicted in Fig.~\ref{page1}(a). 
  However, EDSL directly treats separated symbols as input, which destroys the local continuity between adjacent symbols and causes errors, as the missing exponent symbol depicted in Fig.~\ref{page1}(a). {\bfseries 2). Lack of global context information.} Symbols in expressions are also correlated to the global layout. For instance, the close parenthesis `)' usually appears with the open parenthesis `(', as shown in Fig. 1(b). However, EDSL neglects the context clues of the whole image, which only performs convolution on each local symbol. Even if the multi-head self-attention mechanism between local symbols can capture long-distance relationships, it fails to capture the context cross symbol boundary~\cite{segformer}.
  
  In this paper, we propose a novel Dual Branch Network, named DBN, to fully exploit the context information for accurate mathematical expression recognition. Specifically, DBN adopts a dual branch to simultaneously extract the local context of the symbol and the global context of the whole image. Moreover, we design a Context Coupling Module (CCM) to leverage the complementary advantages between the global and local contexts. CCM adopts an interactive manner so that the coupled context clues are highly correlated to each mathematical expression symbol. With CCM, our network is able to jointly reinforce local and global perception abilities, which is beneficial for recognizing mathematical expressions containing semantically-correlated symbols (e.g., paired braces). Moreover, we argue that there also exists correlations among categories. For example, symbol `c' should be more like `C' rather than `A'. Based on the analysis above, we propose a Dynamic Soft Target (DST) strategy to capture the relationships among target and non-target categories, resulting in more reliable label generation. DST takes advantage of the statistics of model prediction and can update dynamically during the training phase. To summarize, the contributions of this paper are as follows:
  \begin{itemize}
    \item We propose a dual branch network with a Context Coupling Module (CCM) to jointly learn the local and global contexts information for accurate PMER.
    
    \item We design a Dynamic Soft Target (DST) strategy to generate reliable target probability distribution by capturing the similarity among symbol categories.
    
    \item Extensive experiments prove that our proposed DBN has achieved state-of-the-art performance. 
  
  \end{itemize}

\section{Related Work}

The existing PMER methods can be classified into two categories: multi-stage methods and end-to-end methods.

The Multi-stage methods typically include two major steps: symbol recognition phase~\cite{symbolrecoge} and structural analysis phase~\cite{structuralAnalysis}. The symbol recognition phase aims to determine the categories of symbols. The structural analysis is used to parse the symbols into a complete mathematical expression. Okamoto et al.~\cite{structuralAnalysis} analyzed the expression by scanning each component and their relative positions. Moreover, \cite{trees} introduced a structure tree to describe the symbol and recognize expressions using tree transformation. However, multi-stage methods often require predeﬁned expression grammar and suffer from symbol segmentation.

The end-to-end methods~\cite{d3,d4,edsl,WAP} often share an encoder-decoder architecture to recognize the expression. For example, WAP~\cite{WAP} proposed to use RNN equipped with an attention mechanism to generate LaTeX sequences. \cite{Image-to-Markup} introduced a coarse-to-ﬁne attention layer to reduce the inference complexity. \cite{gru1} adopted a GRU network to capture long-distance relationships. \cite{densely} proposed a multi-scale attention mechanism to recognize symbols in different scales. Despite the end-to-end design, the approach will still fall short when facing the mathematical expression containing semantically-correlated symbols.

\begin{figure*}[t!]
  \centering
  \includegraphics[height=3.in,width=4.7in]{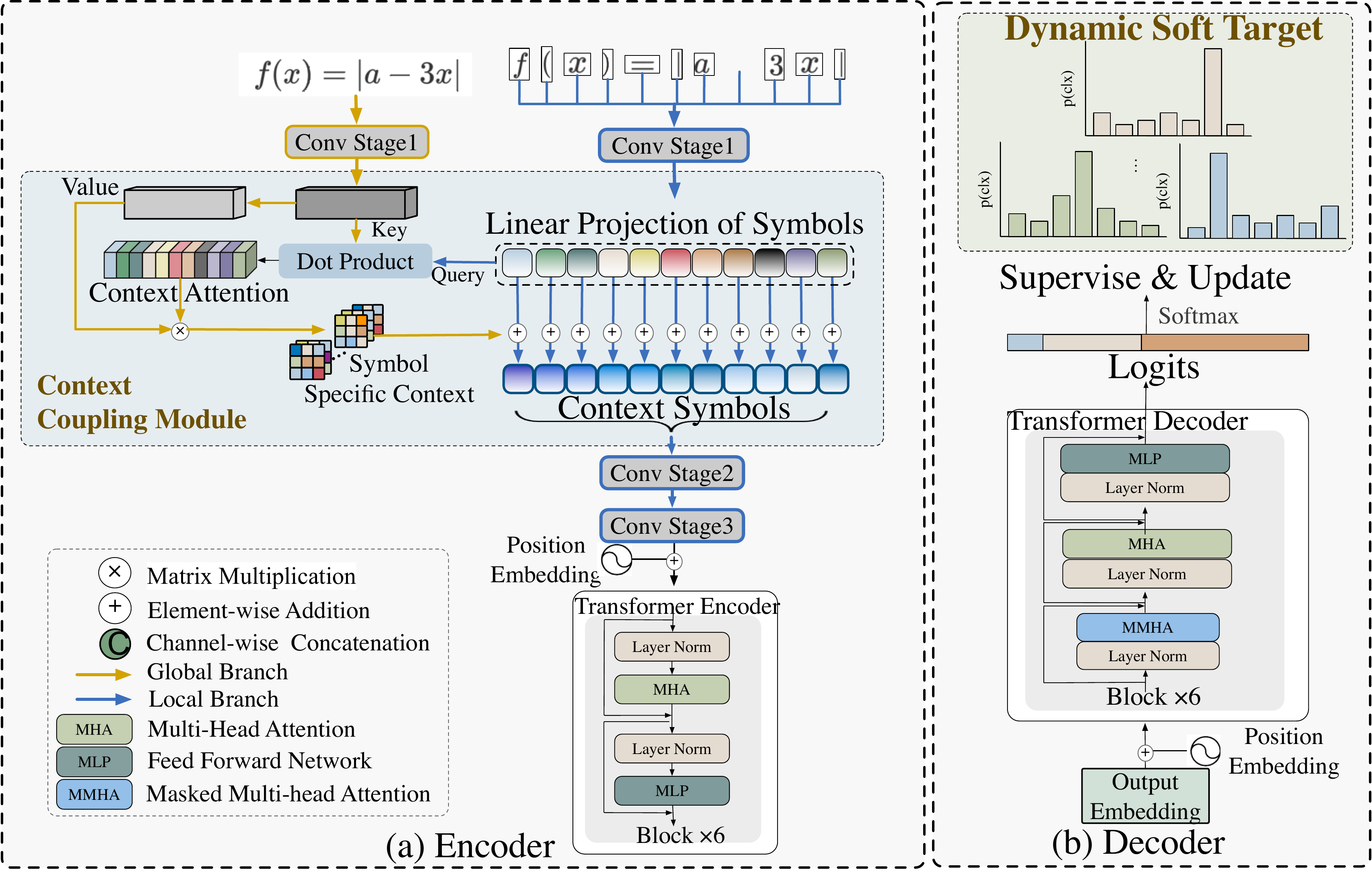}
  \caption{The pipeline of our DBN. The inputs are the whole image and the symbol images segmented from the whole image by the Connected Component Labelling algorithm. Our DBN consists of an (a) encoder to extract image embeddings and a (b) decoder to parse the embeddings into LaTeX sequences. The encoder consists of a dual branch network to jointly learn the global and local context information, a Context Coupling Module (CCM) to couple the context clues for enriching context symbols, then a transformer-based encoder to extract symbol embeddings. The decoder transcribes the embeddings into a LaTeX sequence. Finally, the designed Dynamic Soft Target (DST) strategy updates label distribution for better supervision.}
  \label{pipeline}
\end{figure*}

\begin{figure*}
  \centering
  \includegraphics[height=1.7in,width=4.7in]{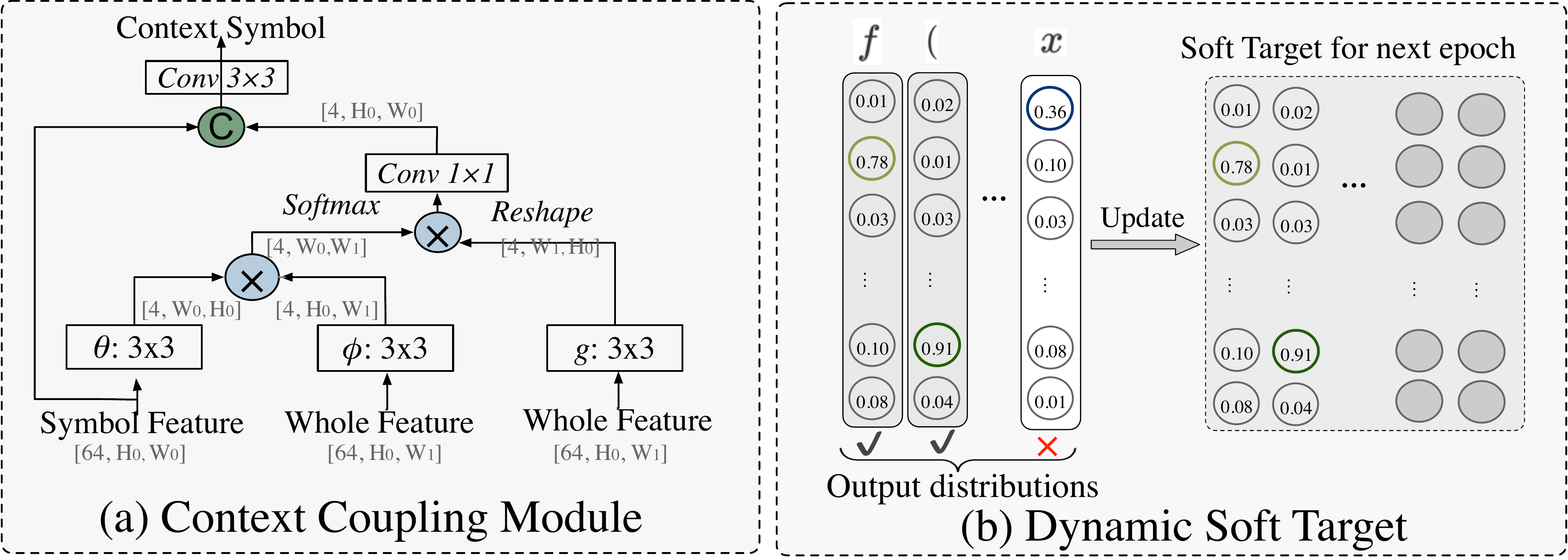}
  \caption{The architecture of our proposed Context Coupling Module(CCM) and Dynamic Soft Target (DST) strategy.}
  \label{ccmdst}
\end{figure*}

\section{Method}
The pipeline of our proposed DBN is illustrated in Fig.~\ref{pipeline}, which follows an encoder-decoder architecture. The encoder includes two different branches: a local context branch extracts feature from local symbols, which are segmented from the given image by Connected-component labeling~\cite{CCL}, and a global context branch extracts feature from the whole image. Then, a Context Coupling Module (CCM) is designed to couple the local and global contexts, as well as produce the enhanced context symbols. After that, the context symbols are sent to a transformer-based encoder to encode symbol tokens. Then, a decoder structure is used to decode the token features into a LaTeX sequence. Finally, the designed DST strategy uses model predictions as input and outputs more reliable label distributions at each training epoch to better supervise the model.

\subsection{Context Coupling Module}
In order to exploit both global image context and local symbol context, we propose a novel Context Coupling Module (CCM) to establish the correlative relationship between local and global contexts.

As illustrated in Fig.~\ref{ccmdst} (a). The input of CCM is the global context extracted by the global branch, which is of shape $ [C, H_0, W_0]$, and the local context extracted by the local branch, which is of shape $[C, H_1, W_1]$, where the $H_0, W_0$ and $H_1, W_1$ are the height and width of the global feature and local feature map, respectively. $C$ indicates the number of channels, set to 8, and the value of $H_0$ and $H_1$ remain the same for further operation. CCM is designed to couple local and global contexts, and it is possible that we directly add the global context to the local context. However, since there is a semantic gap between them, adding them directly will result in feature misalignment. To achieve a better combination, we first align the global features with local features, given by:

\begin{equation}
    \mathbf{y}_{i}=\frac{1}{\mathcal{C}(\mathbf{L})} \sum_{\forall j} f\left(\mathbf{L}_{i}, \mathbf{G}_{j}\right) g\left(\mathbf{G}_{j}\right)^{T}
  \label{equation1}
  \end{equation}
  Here $\mathbf{L}$ is the local context of the symbol, and $\mathbf{G}$ is the global context of the whole image. $g$ computes a representation of the global context. $f$ is a pairwise function that interactively computes the relationship weight between $i$ and $j$. The relationship is used to reweight the global context information for feature alignment. The output feature $y$ is of the same size as $\mathbf{L}$. Finally, $y$ is normalized by a factor $\mathcal{C}(\mathbf{L})$, which is the feature dimension of local context $\mathbf{L}$. In this paper, function $f$ is formulated as:
  \begin{equation}
    f\left(\mathbf{L}_{i}, \mathbf{G}_{j}\right)=Softmax(\theta\left(\mathbf{L}_{i}\right)^{T} \phi\left(\mathbf{G}_{j}\right))
  \end{equation}
  Here, we firstly learn a function $\theta$ that maps $\mathbf{L}$ into new representation, and $\Phi$ maps $\mathbf{G}$ into new representation. Then, we serve local context information as the query, and global information as key, to compute the relationship between local and global representation by matrix multiplication, followed by a Softmax function to normalize them into range $[0,1]$. The relationship can reflect the correlation between the local symbol context and the global image context, ensuring that the aligned global context $y$ is highly correlated to each symbol.

  Finally, we concatenate the aligned global context $y$ with the local context $\mathbf{L}$, and warp CCM into a block operation that can be incorporated into many existing architectures, given by:

\begin{equation}
  \mathbf{z}_{i}=Conv_{1\times1} [\mathbf{L}_{i},\ \mathbf{y}_{i}]
\end{equation}
where $z_i$ represents the output enhanced context symbols. $[\cdot,\cdot]$ denotes channel-wise concatenation, followed by a $1\times1$ convolution layer to improve representation.

\begin{figure*}[t!]
  \centering
  \includegraphics[height=0.9in,width=4.8in]{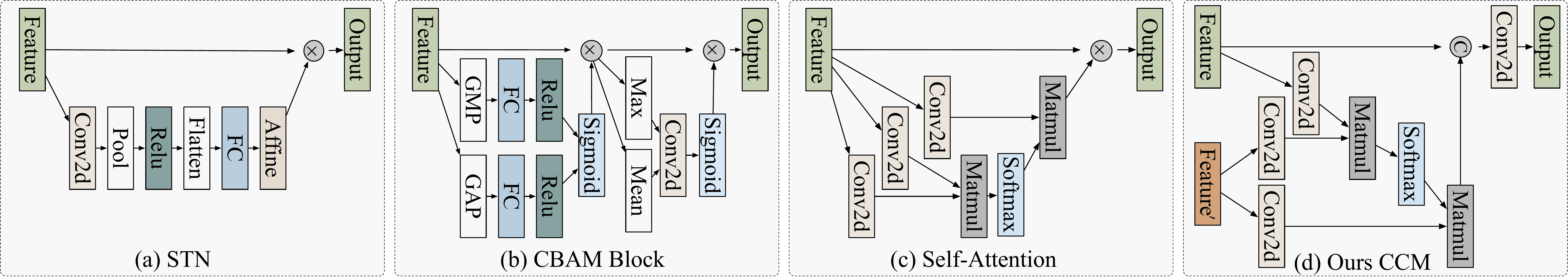}
  
  \caption{Previous solutions for feature enhancement. (a) STN~\cite{STN_jaderberg2015spatial} explicitly predicted important regions. (b) CBAM~\cite{woo2018cbam} implicitly calculates the focus region for feature enhancement. (c) Self-attention mechanism~\cite{non-local}. (d) The proposed CCM builds a bridge between local features and global features for more information propagation. Notes: GAP (global average pooling), GMP (global max pooling), Mean (channel-wise mean operation), Max (channel-wise max operation).}
  \label{attention-vis}
\end{figure*}

Through CCM, we can better align and couple the global image context with the local symbol context, ensuring that more context information gets passed into the model. Moreover, compared with previous solutions in Fig.~\ref{attention-vis} (a-c), which only utilize one branch feature for feature enhancement, our CCM can utilize two branches feature to perceive context information. We also conduct ablation studies to compare our CCM with previous solutions and discuss which convolution stage to add CCM operation in Section~\ref{section_attention}.

\subsection{DST strategy}

Label Smoothing~\cite{LS} is commonly used in PMER, which can generate soft labels by averaging over hard labels and uniform label distributions. Although such soft labels can provide regularization to prevent network over-fitting, it treats the non-target categories equally by assigning them with ﬁxed probability. However, the non-target categories may convey useful information. For example, symbol `c' should be more like `C' rather than `A'. According to knowledge distillation~\cite{2015Distilling}, the model predictions can reflect the similarities between categories.
Inspired by ~\cite{ols}, we use model predictions to capture the relationship among categories and update the soft labels during the training phase, as shown in Fig.~\ref{ccmdst} (b).

Specifically, let $S=\{S^0,S^1,\dots,S^{T-1}\}$ denote the collection of the class-level soft labels, and $T$ represents the number of training epochs. $S^t$ has $K$ rows and $K$ columns, where $K$ is the total number of categories, and each column in $S^t$ corresponds to the soft label for one category. In the $t$ th training epoch, given an input image $\boldsymbol{x_i}$, and its ground truth $y_i=< y_i^1, y_i^2,\dots, y_i^l>$, where $l$ means $l$-th target token in LaTeX sequence. We sum up all the distributions of correctly predicted samples to generate the soft target $S_k^{t-1}$ from $t-1$ epoch:

\begin{equation}
    \left.
      S_k^{t-1} = \frac{1}{C}\sum_{i=1}^N \sum_{j=1}^L p_{i,k}^{j}
    \right.  
 \end{equation}

 where $N$ denotes the number of images in one epoch, and $L$ means the number of correctly predicted tokens for image $\boldsymbol{x}_{i}$ with label $y_{i}^l$. $k$ denotes the target catogery of the current token. $C$ is the normalized factor, which is the total number of correctly predicted samples for one epoch. $p_{i,k}^{j}$ is the output probability of category $k$, given by:
 \begin{equation}
   p_{i,k}^{j}=\left\{
   \begin{array}{ccc}
   p^{t-1}(k|\boldsymbol{x}_{i},y_i^1,y_i^2,...,y_i^{l-1}), &      & if\  \hat{y_{i}^l} = y_i^l\\
   0,    &      & else
   \end{array} \right. 
   \end{equation}

   here, $\hat{y_{i}^l}$ and $y_{i}^l$ is the predicted and ground truth of token $y$ from image $\boldsymbol{x}_{i}$. $p^{t-1}$ is the output probability from the model prediction, given by:
   \begin{equation}
     p=\frac{\exp(z_i/\boldsymbol{T})}{\sum_{j}\exp(z_j/\boldsymbol{T})}
     \label{EquationT}
   \end{equation}
   where $z_i$ represents the output logit from the model prediction, followed by a Softmax function to convert the logit into probabilities. Here, we add a temperature $\boldsymbol{T}$ to smooth the logits. Using a higher value for $\boldsymbol{T}$ produces a softer probability distribution over classes, which contains more knowledge to learn as proved in~\cite{2015Distilling}.
   
   Moreover, in order to fully utilize the soft target generated at each epoch to achieve stable training, we propose to add a factor to adjust the weight between the soft target $S^{t-1}$ from historical epochs and the soft target $S^{t}$ from the current epoch, which is formulated as: 
   
   \begin{equation}
     S^{t}=\beta*S^{t-1}+(1-\beta)*S^{t}
     \label{beta_equation}
   \end{equation}
   where $\beta$ means the reweighting factor, and the larger $\beta$, the more important the distribution in history epochs.

   Finally, the training loss of the model can be calculated by:
   \begin{equation}
     L_{s o f t}=-\sum_{k=1}^{K} S_k^{t-1} \cdot \log p^{t-1}\left(k \mid y_i^1,y_i^2,\dots,y_i^{l-1}\right)
   \end{equation}
   
   The proposed DST indicates that all historically correctly classiﬁed samples $x_n$ will constrain the current sample $x_i$, encouraging samples belonging to the same category to be much closer.

   \section{Experiment}

   Following EDSL, the performance of DBN is evaluated on two benchmark datasets: {\bfseries ME-20K} dataset and {\bfseries ME-98K} dataset. The {\bfseries ME-20K} contains 20,834 expressions from high school math exercises, with a total categories number of 191. 
   The {\bfseries ME-98K} provides a challenging dataset with 98,676 images, which consists of real-world long mathematical expressions collected from published articles~\cite{Image-to-Markup}. The “baseline” mentioned in our experiment refers to the model adapted from EDSL~\cite{edsl}. All ablation studies are conducted on the ME-20K dataset. We use BLEU-4, ROUGE-4, Match, and Match-ws as evaluation metrics in our experiments. Match and Match-ws~\cite{Image-to-Markup} are the exact match accuracy and the exact match accuracy after deleting whitespace columns between the rendered image of ground truth LaTex and predicted LaTex, respectively.

   \begin{table}[h]
    \centering
    
    \caption{Ablation study of the proposed architecture on ME-20K. CCM and DST indicate Context Coupling Module and Dynamic Soft Target strategy, respectively.}
  
    \scalebox{1}{
    \begin{tabular}{p{20pt}<{\centering} p{20pt}<{\centering}|cccc}
    \hline
    CCM & DST  &BLEU-4  &ROUGE-4 & Match & Match-ws  \\
    \hline
    × & ×                 &  94.23 & 95.10 & 92.70 &93.45 \\
    \checkmark   & ×      &94.72& 95.60 & 92.85 & 93.60  \\
    × & \checkmark         &  94.58 & 95.41 & \textbf{92.89} & 93.61 \\
    \checkmark& \checkmark &\textbf{94.73}& \textbf{95.60}&  92.85  & \textbf{93.61} \\
    \hline 
    \end{tabular}}
    \label{ablation}
    \end{table}
\subsection{Implementation Details}
During training, we resize the symbols into size $30\times30$, the same as EDSL~\cite{edsl} for a fair comparison. The embedding size of the transformer is 256. For the proposed CCM, the height of the input image is scaled to 30 while keeping the image aspect ratio. Each convolutional stage in CCM consists of two $3\times3$ convolution layers, followed by a $2\times2$ max-pooling layer to reduce the resolution. We use Adam optimizer with an initial learning rate of 3e-4. For all datasets, the training batch size is 16. The model is trained with 1 Tesla V100 GPU.

    \subsection{Abations Study}
    \label{section_attention}
    As shown in Tab.~\ref{ablation}, we conduct some ablation experiments to illustrate the effectiveness of proposed modules, including the Context Coupling Module (CCM) and the Dynamic Soft Target (DST) Strategy.

    \begin{table}[h]
      \centering
      \caption{Ablation study of one CCM added to different convolutional stages.}
      \begin{tabular}{c|cccc}
      \hline
      Stage & BLEU-4         & ROUGE-4        & Match          & Match-ws       \\ \hline
      1     & \textbf{94.72} & \textbf{95.60} & \textbf{92.85} & 93.60 \\
      2     & 94.64          & 95.50          & 92.70          & 93.33          \\
      3     & 94.34          & 95.14          & 92.84 &  \textbf{93.81}         \\ \hline
      \end{tabular}
      \label{stages}
      \end{table}

      {\bfseries Impact of CCM} is shown in Tab.~\ref{ablation}. We can ﬁnd that it is crucial to couple global and local contexts information. Moreover, we conducted an ablation study to compare our CCM with other feature enhancement modules which only utilize one branch information, and the result is shown in Tab.~\ref{attention_results}. CCM not only boosts the accuracy of baselines signiﬁcantly but also achieves better performance compared with other methods, that is because the powerful feature representation comes from both local and global signals.

      Moreover,  Tab.~\ref{stages} compares a single CCM module added to different convolution stages in Fig.~\ref{pipeline} (a). The block is added right after the last layer of a stage. We can find that CCM is more useful when added to the ﬁrst convolutional stage since it enriches the features. When it is applied to deeper stages, the performance decreases slightly. One possible explanation is that features in deeper stages have a small spatial size due to the pooling layer, and it is insufﬁcient to provide precise spatial information. 

      \begin{table}[h]
        \centering
        \caption{Ablation study of different attention modules.}
        \scalebox{1.0}{
        \begin{tabular}{l|cccc}
        \hline
        Method       & BLEU-4         & ROUGE-4        & Match          & Match-ws       \\ \hline
        Baseline         & 94.23          & 95.10          & 92.70          & 93.45          \\
        w/STN            & 94.28          & 95.27          & 91.93          & 92.85          \\
        w/CBAM            & 94.41          & 95.27          & 92.85          & 93.57          \\
        w/Self-attn       & 94.38          & 95.22          & 92.17          & 92.85          \\
        w/CCM      &\textbf{94.72}& \textbf{95.60} & \textbf{92.85} & \textbf{93.60} \\ \hline
        \end{tabular}}
        \label{attention_results}
        \end{table}

{\bfseries Impact of DST} is shown in Tab.~\ref{ablation}. We can ﬁnd that DST stably boosts the performance of PMER, demonstrating that PMER models can beneﬁt from the soft labels, which brings more knowledge to learn. Using DST to train the model encourages the samples belonging to the same category to be much closer.


\vspace{-5.5 pt}
\subsection{Impact of Hyper-Parameters}
\begin{figure}[h]
  \centering
  \includegraphics[height=1.2in,width=3.1in]{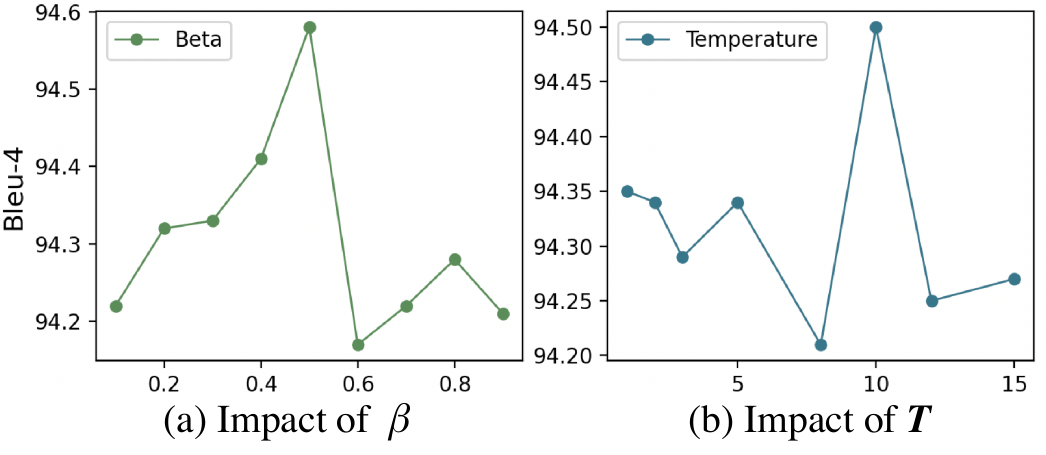}
  
  \caption{Impact of hyper-parameters. The BLEU-4 of different $\beta$ and temperature $\boldsymbol{T}$.}
  \label{temperaturePlot}
\end{figure}

We enumerate possible values with $\beta \in \{ 0.1, 0.2,\dots, 1.0 \}$ for hyperparameter $\beta$ in Eqn.~\ref{beta_equation}. The experiment results are shown in Fig.~\ref{temperaturePlot}(a). It can be seen that the model achieves the highest BLEU-4 score when $\beta$ is set to 0.5. It will decrease the performance when $\beta$ changes from 0.5 to 0. We argue that is caused by the insufficient use of historical labels. When $\beta$ changes from 0.5 to 1.0, the performance decreases sharply. We analyze that the predictions become better and better with the network's training. The accumulated historical labels will be outdated for the model in the current epoch.

Then we analyze the hyperparameter $\boldsymbol{T}$ in Eqn.~\ref{EquationT}. A high value of $\boldsymbol{T}$ softens the probability distribution. It can be seen that the model achieves the highest BLEU-4 score when $\boldsymbol{T}$ is set to 10, as shown in Fig.~\ref{temperaturePlot}(b). We also observe that the performance is very close when $\boldsymbol{T}$ is less than 10. However, when the $\boldsymbol{T}$ is higher than 10, the performance decreases sharply. We argue that this is because that too high $\boldsymbol{T}$ will make the knowledge disappear, as proved in \cite{temperature}.

\begin{table}[h]
  \centering
  \caption{Quantitative results on ME-20K.}
  \scalebox{1.0}{
  \begin{tabular}{l|cccc}
    \hline
    Method & BLEU-4 & ROUGE-4 & Match & Match-ws  \\
    \hline
 DA~\cite{DA} & 87.27& 89.08& 76.92&  77.31 \\
 LBPF~\cite{LBPF}& 88.82& 90.57& 80.46&  80.88\\
 SAT~\cite{SAT}& 89.77& 91.15& 82.09&  82.65\\
 TopDown~\cite{Topdown}& 90.55& 91.94& 83.85&  84.22\\
 ARNet~\cite{ARNet}& 91.18& 92.50& 85.40&  85.84\\
 IM2Markup~\cite{Image-to-Markup}& 92.83& 93.74& 89.23&  89.63\\
 EDSL~\cite{edsl}& 94.23& 95.10& 92.70&  93.45\\
    \hline
     Ours    &\textbf{94.73}& \textbf{95.60}& \textbf{92.85}& \textbf{93.61} \\
    \hline
  \end{tabular}}
  \label{20K}
  \end{table}

  \begin{figure}[t!]
    \centering
    \includegraphics[height=1.8in,width=3.4in]{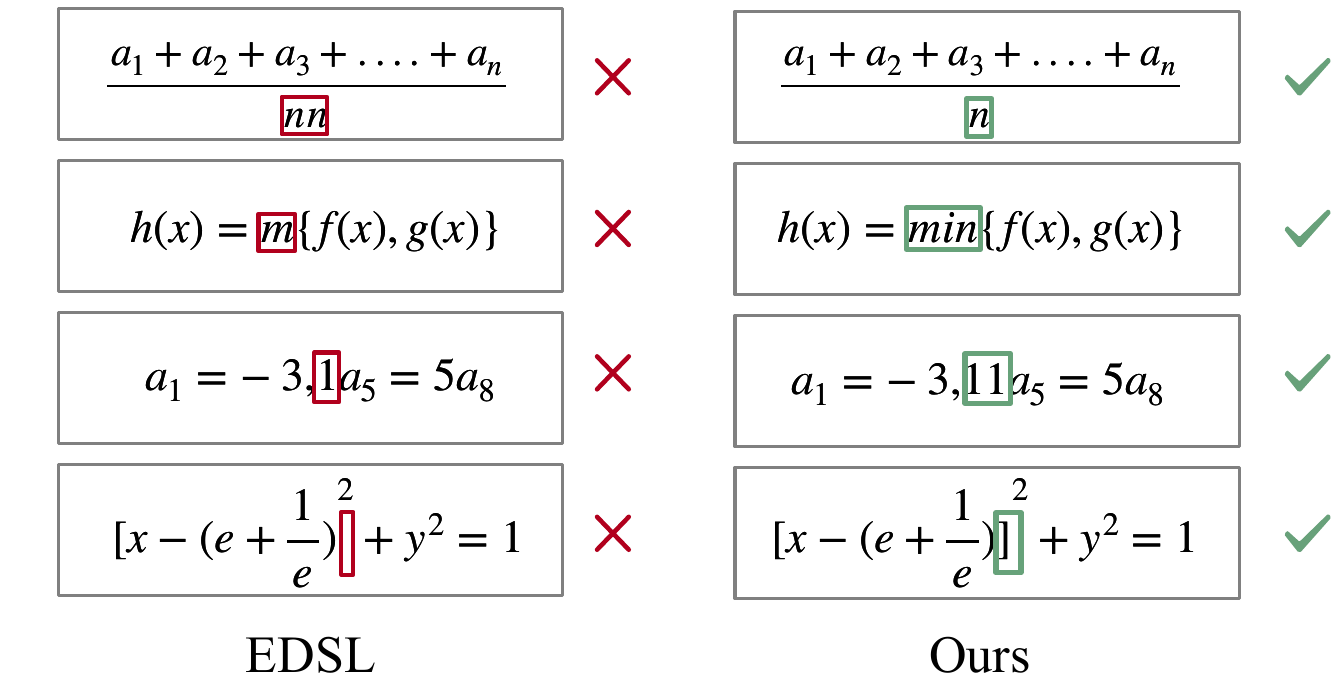}
    
    \caption{Comparisons of our method with EDSL. The red box illustrates the incorrectly predicted symbols from EDSL. The green box illustrates the correct symbols from ours.}
    \label{edsl-ours}
  \end{figure} 

  \begin{table}[h]
   
    \centering
    \caption{Quantitative results on ME-98K.}
    \scalebox{0.9}{
    \begin{tabular}{l|cccc}
      \hline
      Method & BLEU-4 & ROUGE-4 & Match & Match-ws  \\
      \hline
  DA~\cite{DA}&  79.71 &  82.40  & 55.15 &  55.15 \\
  LBPF~\cite{LBPF}&  84.64 &  86.57  & 66.83 &  66.87 \\
  SAT~\cite{SAT}&  86.56 &  87.86  & 70.85 &  71.04 \\
  TopDown~\cite{Topdown}&  87.56 &  89.32  & 72.65 &  72.85 \\
  ARNet~\cite{ARNet}&  86.04 &  88.27  & 68.55 &  68.98 \\
  IM2Markup~\cite{Image-to-Markup}&91.47 &92.45& 84.96 &  85.16 \\
  EDSL~\cite{edsl}&  \textbf{92.93} & 93.30 & 89.00 &  89.34 \\
  \hline
  Ours& 92.90 & \textbf{93.34}&\textbf{89.71}&\textbf{90.01}\\
      \hline
    \end{tabular}}
    \label{98K}
    \end{table}

\vspace{-8pt}
  \subsection{Results on PMER Benchmarks}
  We conduct experiments on 20K and 98K datasets to compare our DBN with other related methods. As shown in Tab.~\ref{20K} and Tab.~\ref{98K}, our proposed method outperforms other methods across the two datasets, which indicates that the proposed method can exploit the inherent context among symbols. 
  The comparison of rendered images in Fig.~\ref{edsl-ours} also shows the priority of our model, which can accurately recognize the expression, while EDSL made some incorrect predictions. Overall, we can prove that jointly coupling global and local contexts is crucial for PMER, and the DST allows samples belonging to the same category to be much closer.

  

  
  \section{Conclusion}
  This paper proposes a novel dual branch framework to jointly learn the local symbol-level and global image-level context for accurate PMER. Moreover, we propose a Context Coupling Module to interactively couple global context with local context so that the obtained context clues are highly correlated to each symbol. Additionally, we propose the DST strategy to capture the similarities among categories by the statistics during the training phase. The quantitative and qualitative experiments have demonstrated that the proposed approach can tackle mathematical expressions with semantically-correlated symbols and has achieved state-of-the-art results.
  
  \section{Acknowledgement}
  This work was supported in part by the National Innovation 2030 Major S\&T Project of China under Grant 2020AAA0104203, and in part by the Nature Science Foundation of China under Grant 62006007.
\bibliographystyle{splncs04}
\bibliography{mybibliography}
%




\end{document}